\documentclass{article}
\usepackage{multirow}
\usepackage{spconf,amsmath,graphicx}
\usepackage{color}
\DeclareMathOperator*{\argmin}{argmin}
\DeclareMathOperator*{\argmax}{argmax}

\title{Domain Adversarial training for accented speech recognition}
\name{{Sining Sun$^{1-3}$\sthanks{Work performed as an intern at Mobvoi AI Lab and University of Washington. 
}, Ching-Feng Yeh$^2$, Mei-Yuh Hwang$^2$, Mari Ostendorf$^3$ , Lei Xie$^1$\sthanks{ Lei Xie is the corresponding author. 
}}}
\address{School of Computer Science, Northwestern Polytechnical University, Xi'an, China$^1$ \\
Mobvoi AI Lab, Seattle, USA$^2$ \\
Department of Electrical Engineering, University of Washington, Seattle , USA$^3$}

\begin{document}
%
\maketitle
\begin{abstract}
In this paper, we propose a domain adversarial training (DAT) algorithm to alleviate the accented speech recognition problem.  In order to  reduce the mismatch between labeled source domain data (``standard'' accent)  
and unlabeled target domain data (with heavy accents), we augment the learning objective for a Kaldi TDNN network with a domain adversarial training (DAT) objective to encourage the model to learn accent-invariant features. In experiments with three Mandarin accents, we show that DAT yields up to $7.45\%$  relative character error rate  reduction when we do not have transcriptions of the accented speech, compared with the baseline trained on standard accent data only. 
We also find a benefit from DAT when used in combination with training from automatic transcriptions on the accented data. 
Furthermore, we find that DAT is  superior to multi-task learning for accented speech recognition. 

\end{abstract}
\ifx\allfiles\undefined 

\section{Introduction}
\label{sec:intro}
There has been significant progress in automatic speech recognition (ASR) due to the development of Deep Learning (DL). DL-HMM based acoustic models are dominating ASR because of their outstanding performance ~\cite{hinton2012deep}. 
However, ASR on speech with background noise, room reverberation, accents, etc. remains difficult even with DL~\cite{45879}. One reason is the mismatch between the training and test data, since it is impossible to cover all kinds of test cases in the training data. In order to alleviate the mismatch, many methods have been proposed in the past from different perspectives such as the front-end signal processing and back-end acoustic modeling. Among these, domain adaptation is also of great interest for  robust speech recognition, especially for DL-based methods.  

Domain adaptation aims for transferring a model trained by the source domain data to the target domain using labeled (supervised) or unlabeled (unsupervised) target domain data. The goal of domain adaptation is to eliminate or reduce the mismatch between the training data and the test data.
Our idea is to learn domain-invariant features to alleviate the mismatch with the help of adversarial training ~\cite{goodfellow2014generative}. Adversarial training has been shown to be successful for domain adaptation problems in the field of computer vision ~\cite{ganin2015unsupervised,tzeng2017adversarial,long2017domain}. Recently, it has been adopted to tackle noise robust speech recognition 
as well ~\cite{shinohara2016adversarial,serdyuk2016invariant,SUN201779}.
In this paper, we focus on {\b unsupervised} accent learning, to minimize expensive and time consuming data labeling efforts.

Our experiments are carried out on large-vocabulary Mandarin speech recognition. Here, the domains we are concerned with are standard Mandarin vs. accented Mandarin.
Our ASR systems are based on the Kaldi Time Delay Neural Network (TDNN) ~\cite{povey2016purely} acoustic model using lattice-free maximum mutual information (MMI) training criterion and the cross-entropy (CE) objective simultaneously, while learning senone posteriors.  We augment the TDNN with another subnetwork to distinguish domain labels (accented vs.\ non-accented), which propagates adversarial signals back to the lower-level shared network to encourage the model to learn domain-invariant features.
Experiments show that DAT can offer up to  $7.45\%$ relative character error rate (CER) reduction. To understand the impact of DAT when used in combination with training using speech transcription, we compare the results with no transcription to performance of systems using ASR-decoded transcription and human transcription. As predicted, performance is the best when human transcription of the target domain data is available. However, in training with ASR transcription, adding the DAT objective continues to have positive (though smaller) impact.
Finally we compare DAT with  multi-task learning (MTL) using a domain classification task, and show that DAT is consistently better than MTL for accented model adaptation.
\section{Related work}
Unsupervised training or adaptation for ASR has been studied for many years. For small amounts of data, such as in speaker adaptation, 
Maximum Likelihood Linear Regression (MLLR) ~\cite{leggetter1995maximum} can be used. Unsupervised training strategies were introduced for leveraging large unlabeled corpora using automatic transcription \cite{Lamel+02,Ma+06,Wang+07}. Later, it was shown that improved results could be obtained using confidence-annotated lattices~\cite{fraga2011lattice}. Other work has looked at the impact of transcription errors and importance sampling using automatic transcription of speech to train deep neural network acoustic models \cite{Huang+16}. In our work, we use the simple 1-best automatically generated transcription, since the focus here is on the interaction with domain adversarial training.

Large scale DL domain adaptation via teacher-student (T/S) learning is proposed to tackle  robust speech speech recognition in ~\cite{li2017large}. In the T/S framework, the source domain data (clean data) comes with human transcription, while the target domain (noisy data) is simulated by adding
various noises to the clean data. The clean data
are first used to train the teacher model. The student model is then trained on the simulated noisy data using the senone posterior probabilities computed by the teacher as soft labels. As it is difficult to generate simulated accented speech, it is difficult to apply this method to the accented speech recognition problem.
Another supervised T/S learning domain adaptation approach is proposed by ~\cite{asami2017domain}. In their work, they combine  knowledge distillation with the T/S model. A temperature T is used to control the class similarity of the teacher model during the process of training.  

Domain adversarial training (DAT) ~\cite{ganin2016domain,ganin2015unsupervised} is also a popular method for DL domain adaptation. Because of its easy implementation and great performance, DAT is commonly used in many computer vision tasks ~\cite{tzeng2017adversarial,zhao2017multiple}. Recently, this method has been applied to noise-robust speech recognition.  In ~\cite{shinohara2016adversarial}, a noise-robust acoustic model is trained using both clean and noisy speech, both with speech transcriptions. At the same time, in order to learn domain-invariant features, an adversarial multi-task is used to predict which domain this frame is from (clean vs.\ a specific noise type). 
Different from~\cite{shinohara2016adversarial}, \cite{SUN201779} applied adversarial training to improve noise robustness in an unsupervised way.

\section{Domain Adversarial training for accented speech recognition}

Accented speech recognition has long been of high interest in industry due to the high recognition error rates. It is difficult to generate simulated accented speech and it is expensive and time-consuming to get plenty of labeled accented speech for training. However, it is relatively easy to collect large amounts of accented speech without transcription. Without loss of generality, we denote the transcribed   standard accent speech data set as $S=\{\boldsymbol{x}_i, y_i\}_{i=1}^{|S|}$, where $\boldsymbol{x}_i$ and $y_i$ are speech and  the corresponding HMM senone labels.  We also have an accented speech data set $T=\{\boldsymbol{x}_i\}_{i=1}^{|T|}$ without transcription. Our goal is to minimize the mismatch between $S$ and $T$ using DAT.

\label{sec:adversarial ASR}
\subsection{Domain invariant features}
\label{subsec:da}
 In our DAT implementation, we pick a layer in the TDNN to represent the domain-invariant feature space. The goal is to learn a feature mapping, $F(\boldsymbol{x})$  to map the input $\boldsymbol{x}$ to a domain-invariant  space $V.$ $V$ yields a distribution $P_V$ and $P(F(\boldsymbol{x}, \boldsymbol{x}\in S))=P(F(\boldsymbol{x}, \boldsymbol{x}\in T))=P_V$. In space $V$, the mismatch between source domain and target domain is reduced, which improves recognition performance on the target domain even when transcribed target data is not available. 
\begin{figure}[htb]
\begin{minipage}[b]{1.0\linewidth}
\centering
\centerline{\includegraphics[width=7cm]{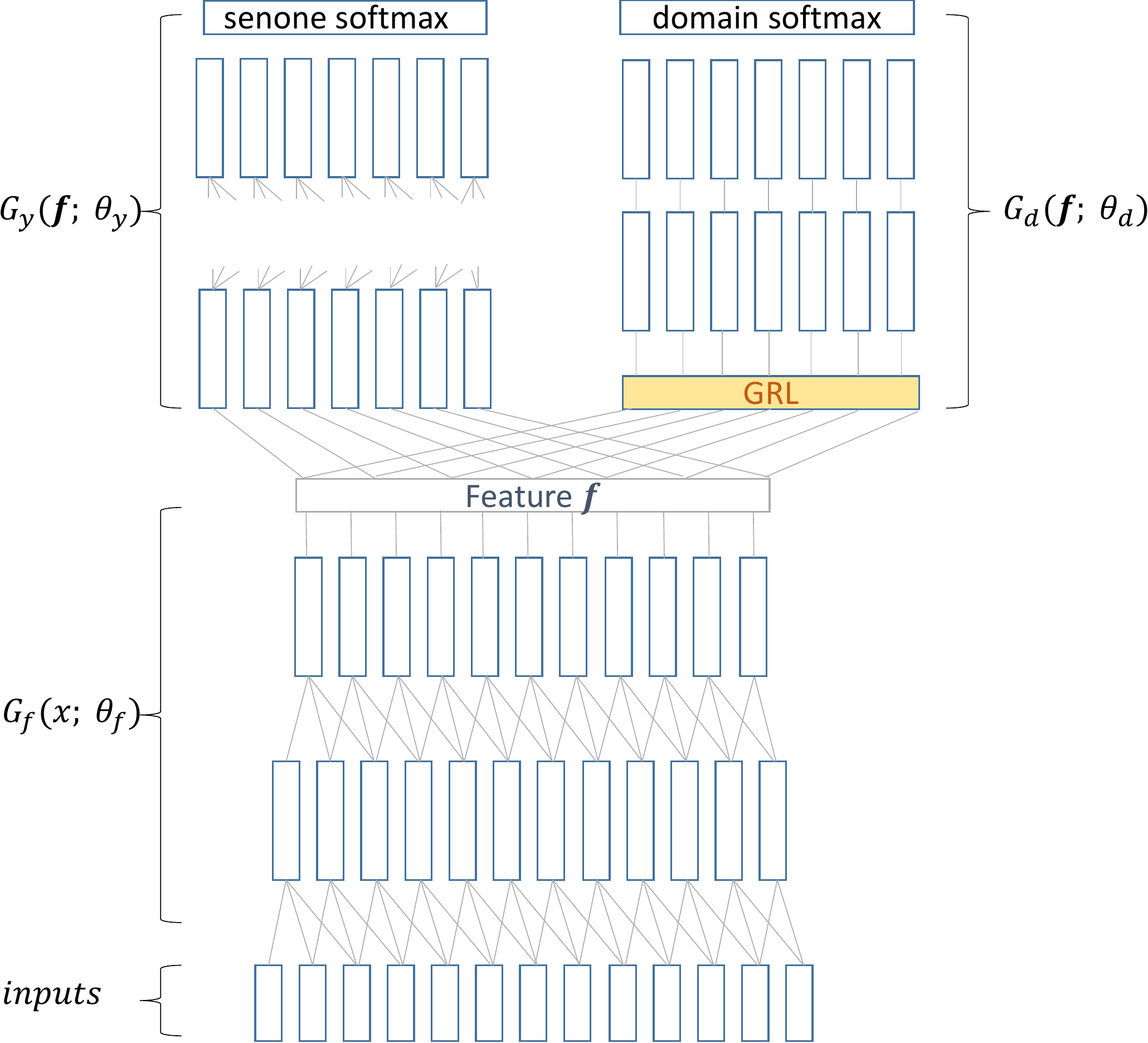}}
  \caption{Domain adversarial training (DAT)}
\label{fig:res}
\end{minipage}
\end{figure}

A typical DAT network is shown in Figure \ref{fig:res}. It consists of three components: the feature generation network $G_f(\boldsymbol{x};{\theta}_f)$ with input speech $\boldsymbol{x}$ and parameters $\theta_f$; the domain classification network $G_d(\boldsymbol{f} ;{\theta}_d)$ with input $\boldsymbol{f}$  and parameters $ {\theta}_d$, which discriminates the source and target domains during the process of training; and the senone classification network $G_y(\boldsymbol{f};{\theta}_y)$ with input $\boldsymbol{f}$  and parameters $ {\theta}_y$. $\boldsymbol{f}$  is the feature generated by $G_f(\boldsymbol{x};{\theta}_f)$ and the goal is to make it invariant to accents. 

\subsection{DAT via back propagation}
\label{DAT}
Assuming there are $N$ frames in a minibatch, the objective function  is:
\begin{equation}
\label{eq1}
\begin{split}
&E({\theta}_f, {\theta}_y, {\theta}_d) = \\
& 
\frac{1}{N}\sum_{i=1}^{N}{(I_d(i)L_{y}^{i} ({\theta}_f, {\theta}_y)} - 
\lambda I_{vad}(i)L_{d}^{i}({\theta}_f, {\theta}_d))
\end{split}
\end{equation}
For DAT, $\lambda$ is a positive hyper parameter. $L_{y}^{i} ({\theta}_f, {\theta}_y)$ is the lattice free MMI loss functions for senone classification network defined in~\cite{povey2016purely}. $L_{d}^{i}({\theta}_f, {\theta}_d)$ is a cross-entropy loss function for the domain classification network, where the target label is binary (accented or not).
 $I_{vad}(i)$ is a voice activity detection (VAD) indicator for training example $\boldsymbol{x}_i$: $I_{vad}(i)=1$ if $\boldsymbol{x}_i$ is speech, otherwise, $I_{vad}(i)=0$. We use the VAD indicator in the loss function, since predicting domain labels for silence segments is nonsense.   $I_d(i)$ is a binary indicator for training example $\boldsymbol{x}_i$, to indicate if this frame is from a transcribed utterance or not. Whenever transcription is available (human or ASR transcription), it is $1$; otherwise it is $0$.
 
The  senone classification network $G_y(\boldsymbol{f};{\theta}_y)$ is optimized by minimizing the senone classification loss, the first item in Equation (\ref{eq1}), with respect to $\theta_y$: 
\begin{eqnarray*}
\theta_y=\argmin_{\theta_y}{ E({\theta}_f, {\theta}_y, {\theta}_d) }.
\end{eqnarray*}
The domain classification network $G_d(\boldsymbol{f};{\theta}_d)$ is optimized by minimizing the domain classification loss 
with respect to 
${\theta}_d$:
\begin{eqnarray*}
\theta_d = \argmax_{\theta_d}{ E({\theta}_f, {\theta}_y, {\theta}_d) }.
\end{eqnarray*}
For the feature generation network $G_f(\boldsymbol{x};{\theta}_f)$, because we want to learn domain invariant features, the feature generated by $G_f(\boldsymbol{x};{\theta}_f)$ should make the well-trained  $G_d(\boldsymbol{f};{\theta}_d)$ fail to distinguish which domain it comes from, and at the same time keep discriminative enough for senone classification. This can be achieved by minimizing the senone classification loss and maximizing the domain classification loss jointly with respect to $\theta_f$. The ``min-max" optimization distinguishes DAT from MTL. When $\lambda > 0$, $\theta_f$ can be optimized by :
\begin{eqnarray*}
\theta_f = \argmin_{\theta_f}{ E({\theta}_f, {\theta}_y, {\theta}_d) }. 
\end{eqnarray*}
That is, while back propagating the error signal
from $G_d(\boldsymbol{f};{\theta}_d)$ to $G_f(\boldsymbol{x};{\theta}_f)$, the bottom layer of the domain classification network acts as a gradient reversal layer (GRL), multiplying the error signal
from the domain classification network by $-\lambda$. On the other hand, if $\lambda <0 $, it becomes a regular multi-task learner.
$\lambda=0$ implies a normal TDNN model. 

To sum up the model parameters are  updated as follows via SGD:
\begin{align}
& \theta_f \leftarrow \theta_f-\alpha \frac{1}{N}\sum_{i=1}^{N}({\frac{\partial{L_y^i}}{\partial{\theta_f}}I_d(i)-\lambda \frac{\partial{L_d^i}}{\partial{\theta_f}}I_{vad}(i))} \\
& \theta_y \leftarrow \theta_y-\alpha \frac{1}{N}\sum_{i=1}^{N}{\frac{\partial{L_y^i}}{\partial{\theta_y}}I_d(i)} \\
& \theta_d \leftarrow \theta_d-\alpha \frac{1}{N}\sum_{i=1}^{N}{ \lambda \frac{\partial{L_d^i}}{\partial{\theta_d}}I_{vad}(i)}
\end{align} 
where $\alpha$ is the learning rate. By adjusting $\lambda$, we can experiment with MTL ($\lambda < 0$), DAT ($\lambda > 0$), or ignore the unlabeled data ($\lambda=0$).

\section{Experiments}
\label{sec:exp}
\subsection{Data}
\begin{table*}
\centering
\caption{Character error rates (CER) of various trainings. 
The baseline system is trained on 360 hours of standard Mandarin (Std). There are 100 hours of training data from each accent. 
With no transcription available on the accented data, we show DAT is effective in learning features invariant to domain differences. }
\label{exp1}
 \scalebox{0.70}[0.8]{
\begin{tabular}{|c|c|c|c|c|c|c|c|c|c|c|c|c|c|c|c|c|c|}
\hline
\multirow{2}{*}{training data} & \multirow{2}{*}{$\lambda$} & \multicolumn{8}{c|}{dev}                                  & \multicolumn{8}{c|}{test}                                 \\ \cline{3-18} 

                        &                            & Std   & FJ    & JS    & JX    & SC    & GD    & HN    & Avg. & Std   & FJ    & JS    & JX    & SC    & GD    & HN    &  Avg. \\ \hline
                        Std                   & -                          & 15.70  & 20.25 & 16.88 & 18.25 & 20.72 & 19.75 & 23.34 & 19.86      & 15.55 & 23.58 & 15.75 & 14.08 & 15.62 & 15.32 & 19.34 & 17.28      \\ \hline
Std + (600hrs with trans)                   & -                          & 14.82  & 10.80 & 10.51 & 11.02 & 11.14 & 13.18 & 15.35 & 12.00      & 14.22 & 14.84 & 9.41 & 8.68 & 9.13 & 9.62 & 11.89 & 10.60      \\ \hline

Std + (600hrs no trans)               & 0.03                       & 15.79 & 19.69 & 16.01 & 17.47 & 20.06 & 19.48 & 21.88 & 19.10     & 15.37 & 22.96 & 14.48 & 13.79 & 15.35 & 14.86 & 18.24 & 16.61     \\ \hline

	\end{tabular}
}
\end{table*}

\begin{table*}[]
\centering
\caption{Results of accent-specific models for accents SC, HN and FJ. All trainings use both Std training data with transcription. Use of accented training data is indicated by MTL ($\lambda=-0.03$), DAT ($\lambda=0.03$) or ``-" if not used.}
\label{tab2}

\scalebox{0.9}[0.9]{
\begin{tabular}{|c|c|c|c|c|c|c|c|c|c|c|c|c|c|}
\hline
\multicolumn{1}{|c|}{\multirow{3}{*}{Accented Data}} & \multicolumn{1}{c|}{\multirow{3}{*}{Training}} & \multicolumn{4}{c|}{SC accent-specific model}        & \multicolumn{4}{c|}{HN accent-specific model}        & \multicolumn{4}{c|}{FJ accent-specific model}         \\ \cline{3-14} 
\multicolumn{1}{|c|}{}                      & \multicolumn{1}{l|}{}                           & \multicolumn{2}{c|}{dev} & \multicolumn{2}{c|}{test} & \multicolumn{2}{c|}{dev} & \multicolumn{2}{c|}{test} & \multicolumn{2}{c|}{dev} & \multicolumn{2}{c|}{test} \\ \cline{3-14} 
\multicolumn{1}{|c|}{}                      & \multicolumn{1}{l|}{}                           & Std    & SC              & Std     & SC              & Std    & HN              & Std     & HN              & Std    & FJ              & Std     & FJ              \\ \hline
\multirow{3}{*}{no trans}                          & MTL                                           & 15.62  & 20.68           & 15.30    & 15.45           & 15.44  & 23.22           & 15.24   & 18.99           & 15.69  & 19.84           & 15.20   & 23.73           \\ \cline{2-14} 
                                            & -                                               & 15.70   & 20.72           & 15.55   & 15.62           & 15.70   & 23.34           & 15.55   & 19.34           & 15.70  & 20.25           & 15.55   & 23.58           \\ \cline{2-14} 
                                            & DAT                                            & 15.44  & \textbf{19.41}  & 15.36   & \textbf{14.72}  & 15.70   & \textbf{21.82}  & 15.16   & \textbf{17.90}   & 15.53  & \textbf{19.09}  & 15.29   & \textbf{22.86}  \\ \hline
\multirow{3}{*}{ASR trans}                          & MTL                                           & 15.85  & 16.05           & 14.74   & 12.15           & 15.50   & 19.30            & 15.25   & 16.19           & 15.40  & 15.17           & 15.35   & 19.68           \\ \cline{2-14} 
                                            & -                                               & 15.63  & 15.77           & 15.38   & 12.05           & 15.59  & 19.82           & 15.13   & 15.81           & 15.32  & 15.19           & 15.13   & 19.27           \\ \cline{2-14} 
                                            & DAT                                            & 15.34  & \textbf{15.62}  & 15.37   & \textbf{11.88}  & 15.52  & \textbf{19.19}  & 15.23   & \textbf{15.62}  & 15.66  & \textbf{15.17}  & 15.45   & \textbf{18.92}  \\ \hline
\multirow{3}{*}{human trans}                          & MTL                                           & 15.05  & 12.83           & 15.08   & 10.45           & 15.33  & 16.99           & 15.22   & 13.58           & 15.26  & 11.72           & 15.32   & 16.54           \\ \cline{2-14} 
                                            & -                                               & 15.32  & 12.79           & 15.37   & \textbf{10.29}  & 15.26  & 16.60           & 14.84   & \textbf{13.52}  & 15.11  & 11.61           & 14.98   & 16.54           \\ \cline{2-14} 
                                            & DAT                                            & 15.50   & \textbf{12.68}  & 14.87   & 10.38           & 15.26  & \textbf{16.21}  & 14.89   & 13.80            & 15.17  & \textbf{11.53}  & 15.04   & \textbf{16.04}  \\ \hline
\end{tabular}
}
\end{table*}

We have about 360 hours of standard accent training data with transcriptions. These data are  voice-search messages from various users. Because they are live logs from various devices and scenarios, they have already covered some background noises and channel variations. Though it likely covers some accented data, most data are relatively standard Mandarin and we name this data set as Std, the source domain set $S$ as introduced in section \ref{sec:adversarial ASR}. The acoustic model trained by the Std set is relatively robust to channel and noise variations but not to accents. Additionally we have a development set (dev) and a test set (test) for standard Mandarin speech, each with 2000 sentences.

We purchased 100 hours of Mandarin speech per accent from 6 different provinces in China. These accents are: HuNan (HN), SiChuan (SC), GuangDong (GD), JiangXi (JX), JiangSu (JS) and FuJian (FJ). These data come with human transcriptions. However, we will use this data set as the target domain data set $T$, as though the transcriptions were not
available. For each accent, there are separate dev and test sets, each with 2000 sentences. 

In reporting CER, we use the dev set to find the optimal language model weight, and then apply the best language weight to the test set.

\subsection{Invariant feature extraction  across all accents}
Our baseline TDNN acoustic model (Row 1 in Table \ref{exp1})  is trained using 360 hours of Std data without domain adversarial training. This Std baseline consists of 7 layers and each layer has 625 hidden units with ReLU activation functions and 5998 softmax output units. We use 23-dimensional filterbanks with 3 pitch features as our acoustic feature vector. Three consecutive frames are concatenated as the input to the TDNN. The acoustic model is trained by Kaldi~\cite{povey2011kaldi} using the criteria proposed by~\cite{povey2016purely}, with a subsampling rate of 3, both at training and decoding time. All experiments share the same network configuration as Std. 
Comparing the different results in row 1 to the Std case shows the performance degradation due to domain mismatch. The second row shows the gains possible when hand-transcribed multi-domain training data is available. 

Next assuming the transcription of the accented speech is not available, we explore how much performance can be improved using only the knowledge of the accent class in DAT, via the domain classifier $G_d(\boldsymbol{f};{\theta}_d)$.
In this experiment, we use all Std data and all 600 hours of accented data without transcriptions to train the model.  There are two hidden layers in the domain classifier network, where each layer has 625 ReLU units. The input of the domain classifier is the activation of the second hidden layer of the baseline Std network. 
 With the domain classifier, we tried a few $\lambda$'s and the best result is from $\lambda=0.03$ shown in the third row of Table \ref{exp1}, indicating the 
 effectiveness of adversarial learning without compromising recognition on Std speech.
 
 It shows that DAT can help with all types of accents even without human transcriptions on the accented training data. 
The average error reduction across the different accents is 3.8\%.

\subsection{Accent-specific adversarial training}

We are also interested in accent-specific adaptation, where the Std model is adapted per accent.  Compared with multiple accents, the single accent variance is smaller and thus we expect to get better results. 
Three accented data sets, FJ, SC and HN, are 
selected to do accent-specific experiments, based on the highest baseline CER on the dev set. We  investigate three cases: 1) no transcriptions of accented data are available,  2) approximate transcriptions of the accented training sets are obtained by decoding them using the baseline Std acoustic model, and 3) human transcriptions of the accented training data are available. We compare DAT ($\lambda>0$) with MTL ($\lambda<0$), and no use of unlabeled data ($\lambda = 0$).

From Table \ref{tab2} we can see that DAT is always helpful in  dev and test sets in the first two cases, when the correct transcription is not available. The performance of multi-task learning is inconsistent, where sometimes it helps a little but more often it hurts the accuracy. This is because multi-task optimization is learning domain-discriminative features, which can be at odds with the senone classification task.
In contrast, DAT can learn more accent-invariant features, especially when we cannot access the true labels of the target domain data. When no transcription on the accented data is available, DAT gave 
$5.8\%$, $7.4\%$, and $3.1\%$ 
relative CER reduction in SC, HN and FJ accent respectively, compared with the Std model.

When unsupervised or supervised transcription becomes available, the DAT contribution shrinks. With more detailed knowledge about the target data, 
the unsupervised DAT becomes less important. 

\section{Conclusion}
\label{sec:conc}

In this paper, we integrated unsupervised domain adversarial training (DAT) into TDNN acoustic model training to tackle the accented speech recognition problem. We compared DAT with MTL in different setups 
and observed that DAT was more effective for different transcription scenarios and different domains.
Compared with the model trained on standard accent data exclusively, DAT with a binary domain label provided 
up to $7.4\%$ relative CER reduction. Combining DAT with unsupervised adaptation via automatic transcription of the accent data gives an overall CER reduction of 20\%. 

The concept of DAT is not limited to adapting to accented speech only. As noted earlier, it has been successfully applied in other scenarios such as learning channel-invariant features for robustness in different recording conditions. In the future, we will explore the possibility of applying it to far-field speech recognition.

\section{Acknowledgement}
\label{sec:ack}
The research work is supported by the National Key Research and Development Program of China (Grant No. 2017YFB1002102) and the National Natural Science Foundation of China (Grant No. 61571363).Thanks Yating Wu for her useful comments on drafts of this paper.

\vfill\pagebreak

\bibliographystyle{IEEEbib}
\bibliography{main}

\end{document}